ORIGINAL MANUSCRIPT

# A new classification system of beer categories and styles based on large-scale data mining and self-organizing maps of beer recipes

Diego Bonatto[a,*]

[a]Laboratório de Biologia Computacional e Molecular, Centro de Biotecnologia da Universidade Federal do Rio Grande do Sul, Departamento de Biologia Molecular e Biotecnologia, Universidade Federal do Rio Grande do Sul, Porto Alegre, RS, Brazil

**Short title:** Dataming and Beer Categories Analysis

**\*Corresponding author:**
Diego Bonatto
Centro de Biotecnologia da UFRGS - Lab. 212
Departamento de Biologia Molecular e Biotecnologia
Universidade Federal do Rio Grande do Sul – UFRGS
Avenida Bento Gonçalves 9500 - Prédio 43431
Caixa Postal 15005
Porto Alegre – Rio Grande do Sul
BRAZIL
91509-900
Phone: (+55 51) 3308-9410
E-mail: diego.bonatto@ufrgs.br
Contract/grant sponsor: CNPq

**Abstract**

A data-driven quantitative approach was used to develop a novel classification system for beer categories and styles. Sixty-two thousand one hundred twenty-one beer recipes were mined and analyzed, considering ingredient profiles, fermentation parameters, and recipe vital statistics. Statistical analyses combined with self-organizing maps (SOMs) identified four major superclusters that showed distinctive malt and hop usage patterns, style characteristics, and historical brewing traditions. Cold fermented styles showed a conservative grain and hop composition, whereas hot fermented beers exhibited high heterogeneity, reflecting regional preferences and innovation. This new taxonomy offers a reproducible and objective framework beyond traditional sensory-based classifications, providing brewers, researchers, and educators with a scalable tool for recipe analysis and beer development. The findings in this work provide an understanding of beer diversity and open avenues for linking ingredient usage with fermentation profiles and flavor outcomes.

# 1. Introduction

Beer is one of the oldest and most culturally significant fermented beverages, with a documented history dating back over 6,000 years to ancient Mesopotamia and Egypt. Its production has evolved over centuries, influenced by regional traditions, raw material availability, technological innovation, and shifting consumer preferences (Hornsey, 2003). As a result, beer styles have proliferated, encompassing various sensory profiles, alcohol contents, ingredients, and fermentation techniques (Baiano, 2021). Systems such as the Beer Judge Certification Program (BJCP) and Brewers Association guidelines have historically been used to classify beer styles based on sensory attributes, geographic origin, and traditional recipes to organize this diversity. Although these systems are valuable for training sensory panels, guiding competition, and educating brewers, they remain inherently subjective. Classifications often depend on descriptive terms such as "malty," "hoppy," or "balanced," which are influenced by individual perception and cultural context (Villacreces et al., 2022). Moreover, they rarely incorporate quantitative data regarding ingredients, fermentation metrics, or biochemical outputs. With the rise of digital brewing platforms and the global craft beer movement, brewers can access thousands of publicly available recipes, offering a unique opportunity to reclassify beer styles based on objective data and statistical analysis.

In this context, data-mining approaches and machine-learning techniques can offer new perspectives for beer classification by identifying latent patterns in recipe composition and brewing practices. Self-organizing maps (SOMs) represent a robust unsupervised learning method capable of reducing high-dimensional data while preserving topological relationships among input variables (Kohonen, 1990). SOMs have been successfully employed in food science to cluster sensory and compositional data; however, their application in brewing science remains underexplored. The present study aimed to develop a comprehensive and reproducible classification of beer styles by applying data mining tools to over 62,000 beer recipes. Recipes were filtered to ensure the completeness of key parameters, including the number and types of malts, hops, adjuncts, and fermentation indicators, such as original gravity, final gravity, alcohol by volume (ABV), and international bitterness units (IBU). These data generated a dissimilarity matrix using Gower's metric, incorporating numerical and categorical features. The matrix was used to train the SOMs, which grouped beer styles into clusters and superclusters based on their shared characteristics. The findings identified four major superclusters that partially align with traditional classifications but reveal hidden relationships driven by ingredient usage and brewing methodology. For instance, while conventional lager styles are clustered together based on restrained hopping and conservative grain bills, many hot-fermented beers form distinct groups based on the complexity of malts, use of adjuncts, or hopping techniques such as dry hopping and whirlpool additions. This study bridges the gap between traditional sensory-based classification systems and modern data-driven approaches by offering a quantitative and scalable framework for beer style classification. This provides new tools for brewers, researchers, and educators to navigate the rapidly expanding landscape of beer styles and brewing techniques. Moreover, the proposed classification supports future studies linking brewing inputs to biochemical, sensory, and consumer response outputs, ultimately contributing to scientific understanding and innovation.

## 2. Materials and methods

*2.1. Beer recipe data prospection and analysis*

Data regarding beer recipes were obtained from Brewer's Friend (https://www.brewersfriend.com; last accessed from November 2020 to December 2022) with the direct consent of the webpage administrator. Initially, the Lynx web browser (https://lynx.browser.org) was used to map all links associated with beer recipes, beer categories, and styles from Brewer's Friend. Once obtained, the library vest (https://github.com/tidyverse/rvest) from the R software (https://www.r-project.org) scraped recipe information for different beer categories and styles from Brewer's Friend links. The raw recipe data obtained were filtered considering their completeness in terms of malt, hops, adjunct types and quantities, recipe vital statistics [alcohol by volume (ABV), beer color in standard reference method (SRM), standard gravity (SG), final gravity (FG), international bitterness units (IBUs), mashing procedures, and hop usage]. The resulting filtered recipe data were merged with the beer category information (Table S1), where each beer category is a large group of beer subcategories and/or styles that usually have similar sensorial characteristics and/or share similar ingredients and/or geographical regions where they are produced. The definitions of beer categories (Table S2), as well as the country or geographical area from which they originated, were obtained from the 2021 Beer Judge Certification Program (BJCP) Style Guidelines (https://dev.bjcp.org).

*2.2. Statistical and quantitative data analysis*

The R software (https://www.r-project.org) was used for statistical and quantitative data analyses. Data normality for quantitative variables for each beer category and style was evaluated using a univariate Shapiro-Wilk normality test, followed by analysis of homogeneity of variance (Levene's test) implemented in the rstatix library (https://cran.r-project.org/web/packages/rstatix/index.html). The average number of different hops, grains (malts), dry and liquid malt extracts, adjuncts (non-malted grains or grain husks), sugar, and fruits employed in cold or hot fermented beers were statistically evaluated using the Bootstrap-t method for a one-sample test implemented in the ggstatsplot library (Patil, 2021) with the following parameters: display confidence interval ($CI_{95\%}$) and trimmed average value (µ), and evaluation of the exact p-value. The number of different hops, grains (malts), dry and liquid malt extracts, adjuncts (non-malted grains or grain husks), sugar, and fruits used for cold or hot fermented beers were evaluated by parametric (Welch's t-test) or non-parametric (Mann-Whitney test) analysis implemented in ggstatsplot (Patil, 2021) and represented as a heatmap using the ComplexHeatmap library for R (Gu, 2022; Gu et al., 2016). The library seriation for R (Hahsler et al., 2008) was employed for heatmap data ordering using hierarchical clustering with the optimal leaf ordering (OLO) method (Bar-Joseph et al., 2001). Different malt types (base, crystal, roasted, specialty, acidulated, smoked, and gluten-free malts; Table S3) contribution to beer categories and styles (denominated as "malt preferential usage") was evaluated using the percentize method implemented in the library heatmaply for R (Galili et al., 2017). Moreover, the average number of distinct malt types employed in different beer categories and recipes was initially determined by considering the relationship among the following sets: $R(c)$, the set of beer recipes associated with beer category $c$; $M(c, r)$, the set of malt types used in beer recipe $r$ of beer category $c$; and $S(c, r, m)$, the set of malt subtypes associated with malt type $m$ in beer recipe $r$ of beer category $c$. To determine the number of distinct malt subtypes $s$ for each beer category $c$, beer recipe $r$, and malt type $m$, the algebraic expression below was employed (Equation 1):

$$T(c, r, m) = |S(c, r, m)| \quad \text{(Equation 1)}$$

Then, the number of distinct malt types by beer category was calculated considering all recipes $r$ and all malt types $m$ associated with those recipes by employing a function $F(c)$ that aggregates the distinct malt types $m$ used in all recipes $r$ for each beer category $c$ (Equations 2 and 3):

$$F(c) = \left| \bigcup_{r \in R(c)} \bigcup_{m \in M(c, r)} \{m\} \right| \quad \text{(Equation 2)}$$

$$F(c) = |\{m \in M \mid \exists r \in R(c), \exists s \in S(c, r, m)\}| \quad \text{(Equation 3)}$$

Finally, the average number of malt types employed by recipe $A(c)$ was determined as follows (Equation 4):

$$A(c) = \frac{|M(c)|}{|R(c)|} \quad \text{(Equation 4)}$$

Where $|M(c)|$ is the number of specific malt types in the beer category $c$, and $|R(c)|$ is the number of recipes in the beer category $c$.

To evaluate the average number of malt subtypes employed in a beer style $i$ and malt types $k$, the following equation was employed (Equation 5):

$$A_{sm} = \frac{\sum_r D_{sr}^m}{Rm} \quad \text{(Equation 5)}$$

where $A_{sm}$ is the average number of distinct malt subtypes for beer style $s$ and malt type $m$; $D_{sr}^{m}$ is the number of distinct malt subtypes for beer style $s$, beer recipe $r$, and malt type $m$; and $R_s$ is the total number of beer recipes for beer style $s$.

To calculate the percentage (or ratio) of a specific malt type used within a particular beer category, the following algebraic equation was used (Equation 6):

$$P = \frac{M_{specific}}{M_{total}} \times 100 \text{ (Equation 6)}$$

where $P$ is the percentage of the specific malt type used in the beer category, $M_{specific}$ is the total mass of the specific malt type used in the beer category, and $M_{total}$ is the total mass of all the malt grains used in the beer category.

The preferential usage of different beer hopping methods (boil, aroma, dry hop, dry hop at high krausen, whirlpool, first wort, hop stand, hopback, and mash) regarding beer recipes was determined using the percentile method implemented in a heatmaply library for R (Galili et al., 2017). The relative bitterness ratio (RBR) of different beer categories was calculated as follows (Equation 7):

$$RBR = \left(\frac{IBU}{SG}\right) \times (1 + (ADF - 0.7655)) \text{ (Equation 7)}$$

where $IBU$ stands for "international bitterness units," which represents the concentration of isomerized alpha acids (mg/mL), $SG$ (wort standard gravity), and $ADF$ (apparent degree of fermentation).

The contribution of each beer hopping method to the IBUs of different beer recipes belonging to a specific beer category (average IBU) was defined using the following data sets: $C$ (set of beer categories), $Rc$ (set of beer categories within a category $c \in C$), $Hrc$ (set of beer hopping methods within a recipe $r \in Rc$), and IBU$hrc$ (IBU value for a hopping method $h \in Hrc$). The average (mean) IBU value for each hopping method within each beer recipe is defined as follows (Equation 8):

$$\widehat{IBUhrc} = \frac{1}{|Hrc|} \sum_{h \in Hrc} IBUhrc \text{ (Equation 8)}$$

By its turn, the average IBU value for each beer recipe within each beer category was determined with the following equation (Equation 9):

$$\widehat{IBUrc} = \frac{1}{|Rc|} \sum_{r \in Rc} \widehat{IBUhrc} \text{ (Equation 9)}$$

Finally, the average IBU value for each beer category was estimated by applying the equation below (equation 10):

$$\widehat{IBUc} = \frac{1}{|C|} \sum_{c \in C} \widehat{IBUrc} \text{ (Equation 10)}$$

A nested summation of IBUs over hopping methods and recipes for each beer category (Equations 8–10) gave the average IBU value by beer category ($\widehat{IBUc}$; Equation 11):

$$\widehat{IBUc} = \frac{1}{|Rc|} \sum_{r \in Rc} \left( \frac{1}{|Hrc|} \sum_{h \in Hrc} IBUhrc \right) \text{ (Equation 11)}$$

The average IBU value by beer category was represented as a heatmap using the library ComplexHeatmap for R (Gu, 2022; Gu et al., 2016), followed by heatmap data ordering using the OLO method (Bar-Joseph et al., 2001) implemented in the library series for R (Hahsler et al., 2008).

The average number of distinct hops employed by different beer styles $i$ and beer hopping methods $k$ was defined by the following equation (Equation 12):

$$A_{ik} = \frac{\sum_{j} D_{ij}^{k}}{Ri} \text{ (Equation 12)}$$

where $A_{ik}$ is the average number of distinct hops for beer style $i$ and hopping method $k$; $D_{ij}^{k}$ is the number of distinct hops for beer style $i$, beer recipe $j$, and hopping method $k$; and $R_i$ is the total number of beer recipes for beer style $i$.

The cumulative percentage, implemented in R software (https://www.r-project.org), was employed to evaluate the trend in using different malt and hop techniques for each beer style with a cutoff ⩾ of 50%.

### 2.3. Self-organizing maps of beer styles

The self-organizing map (SOM) technique (Kohonen, 1990) was employed to classify beer styles into superclusters and clusters. Initially, a data frame was generated using the average number of different malt subtypes and hops employed by different beer styles, malt types, and beer-hopping methods (Equations 6 and 12, respectively) (Table S4). Additionally, the average values of SG, final gravity (FG), ADF, ABV, SRM, and IBU for each beer style were included in the dissimilarity matrix (Table S4). The library cluster for R (Maechler et al., 2023) was employed to obtain a dissimilarity matrix from the data frame generated above with the following options: daisy algorithm (Kaufman & Rousseeuw, 1990), and Gower's metric. Once generated, the dissimilarity matrix was employed in the R library SOMbrero (Olteanu & Villa-Vialaneix, 2015) to train the artificial neuron network. The output defined the major superclasses and classes associated with beer style (Table S5).

### 3. Results and Discussion

### 3.1. Beer recipes and categories data analysis

By applying data mining tools for beer recipe prospection, it was possible to download 114,347 recipes from the Brewer's Friend website (https://www.brewersfriend.com). However, many beer recipe entries were incomplete in describing the major components of beer production, such as malt types/subtypes, hops, adjuncts, and sugars. Additionally, many entries lack recipe vital statistics, a key numerical element in beer production that describes the essential characteristics of a beer (Palmer, 2017), including measurements such as alcohol by volume (ABV), beer color in standard reference method (SRM) or any other established method (e.g., European beer convention method or EBC), standard gravity (SG), final gravity (FG), and international bitterness units (IBUs). Moreover, the beer recipes were filtered against incompleteness regarding mashing procedures and hop usage. Thus, approximately 45.67% of the beer recipes were discarded because of their incompleteness, resulting in 62,121 recipes employed for subsequent analyses (Table S1). From this sample, 5,130 recipes were cold-fermented beers, whereas 56,991 were hot-fermented beer recipes (Table S1) (Bonatto, 2021).

An essential aspect of any beer recipe is the description of the major components/ingredients employed for its manufacturing (Baiano, 2021), which are specified by the malt types and subtypes (indicated as "grains" in this work), hops, dry and liquid malt extracts, adjuncts (mainly non-malted grains and husks), sugars (e.g., sucrose, lactose, honey, among others, and its different formulations, like raw, processed, caramelized, among others), and fruits. Different beer schools worldwide employ components/ingredients in beer manufacturing according to traditional historical recipes or by attending to innovative demands promoted by local consumers or tourists (Baiano, 2021; Postigo et al., 2024; Villacreces et al., 2022). Therefore, considering the filtered databank of beer recipes (Table S1), an analysis of the most predominant components/ingredients in cold- and hot-fermented beers was performed (Figures 1A and B). The data gathered point to the use of different types of grains and hops in all beer recipes (Figures 1A and B), where cold- and hot-fermented beers employed at least three different grain types (a combination of base, crystal, roasted, specialty, acidulated, smoked, and gluten-free grains) for beer recipe formulation (Figures 1A and B). Interestingly, approximately 83.3% of the beer recipes employed at least one to two different hops for cold-fermented beer (Figure 1A), while hot-fermented beer employed at least two to three different hops (Figure 1B). Considering other major components/ingredients for beer recipe formulation, such as dry and liquid malt extracts, adjuncts, sugars, and fruits, no statistical significance was observed among the number of components employed in the cold- ($p$ = 0.04) and hot-fermented ($p$ = 0.13) beers for a trimmed average value of 1.22 different components (Figures 1A and B).

Counting unique beer ingredients in cold- and hot-fermented beer indicated that many different grains and hops were employed for different beer categories. For example, the India Pale Ale category (IPA, Figure 2A) employs 180 different grains and 226 hop types, and a similar result was observed for Pale American Ale (Figure 2A). Another example is the Fruit Beer category, which uses 35 fruits (Figure 2A). In contrast to hot-fermented beers, cold-fermented beers employ significantly fewer grains, hops, adjuncts, fruits, and sugars (Figure 2B), indicating a conservative view of brewers concerning the design of cold-fermenting beer recipes and the use of innovative ingredients. No statistical differences were observed between the dry and liquid extracts of the cold- and hot-fermented beers (Figure 2B). Because grains and hops are the most employed components in all beer categories (Figures 1 and 2), how these components/ingredients are used and their impact on different beer categories were evaluated.

*3.2. Analysis of grain types employed in different beer categories*

In brewing, the use of different malt types and their subtypes contributes to the flavor, aroma, and color of wort/beer, in addition to being a source of free amino nitrogen, simple and complex carbohydrates, lipids, fatty acids, nitrogen bases, and micronutrients that are used by yeast during fermentation (Briggs, 1998; Castro et al., 2021). Moreover, malt also contributes to enzymes during the wort production phase, especially diastases ($\alpha$-amilase and $\beta$-amilase) and proteases, which convert the starch and proteins from malt endosperm to simple carbohydrates, peptides, and amino acids (Guido & Ferreira, 2023; Jones, 2005). Barley is the major grain employed for malting, but others can be malted and used in brewing, like wheat, oats, rye, and sorghum (Blšáková et al., 2022; Hager et al., 2014).

Generally, barley malt is classified into different types that are defined according to the green malt (germinated seed) dry process (kilning) and roasting temperatures, which affect both color and flavor (Castro et al., 2021; Guido & Ferreira, 2023). In this sense, the base malt and its subtypes (e.g., Pilsner malt and Pale Ale malt) are employed in all beer categories and styles because of their high enzymatic activity, favoring the conversion of malt starch and proteins to simple compounds (Prado et al., 2021). Specialty malts (e.g., Vienna malt and Munich malt) are produced by submitting green malt to high kilning and roasting temperatures, promoting the formation of Maillard compounds, diminishing its enzymatic activity, and promoting color and flavor for specific beer categories and styles (Guido & Ferreira, 2023). For roasted malts and subtypes (e.g., Biscuit, Chocolate, Black Patent malts), the green malt is subjected to high roasting temperatures to promote the formation of pyrazines, pyridines, and pyrroles that are associated with the characteristic flavor of these malts, and the roasting temperature promotes the formation of complex molecules related to dark colors (Carvalho et al., 2016; Castro et al., 2021). Crystal/caramel malts (e.g., Carapils, Crystal 30L, Crystal 45 L) are produced from un-kilned green malt, where the grains are heated to $\alpha$- and $\beta$-amylase optimal temperature to induce the endosperm mashing inside the husk followed by grain exposition to high temperature to promote the formation of flavors associated with caramel and sugar (Castro et al., 2021). The other two malt types are employed to a lesser extent in brewing: acidulated/sour malt and smoke malt. Acidulated/sour malt is produced by incubating malt in water at 45 to 48 °C to promote the formation of lactic acid by lactic acid-producing bacteria (LAB), followed by drying at low temperature (Prado et al., 2021); finally, smoked malt is produced by exposing green malt to a smoking system that impregnates the malt with volatiles containing phenolic compounds like guaiacol and eugenol from burning substances (e.g., wood or peat), (Prado et al., 2021). Besides traditional malt types, gluten-free malts are gaining attention due to their application for producing gluten-free beer (Hager et al., 2014), a high-value, small-scale niche market that attends consumers who suffer from celiac disease or search for healthy products (Harasym & Podeszwa, 2015). Gluten-free cereals (e.g., oats, rice, and corn) and pseudocereals (e.g., buckwheat, quinoa, and amaranth) can be malted and used instead of traditional barley, wheat, and rye malts, despite their low extract yield in wort (Yang & Gao, 2022). Thus, it is essential to understand the relationship among beer categories and styles to evaluate how brewers employ these different malt types and subtypes in their recipes.

The data gathered from beer recipes regarding the use of malt types in different beer categories showed that two major groups of malt types were identified, considering the average number of grains by recipe and the percentage of grist composition (Figure 3). The first group comprises malt types found in specific beer categories, such as acidulated and smoked malt types, which are employed in sour and smoked beers, respectively (Prado et al., 2021). In contrast, gluten-free malts mainly compose gluten-free beer from different categories (Yang & Gao, 2022). This very narrow usability of acidulated, smoked, and gluten-free malts is translated by the low average number of different grains employed (Figure 3), as indicated for acidulated ($M = 0.05$ grains), smoked ($M = 0.008$), and gluten-free malts ($M = 0.002$). Looking at the grist composition, independent of beer categories, acidulated malt comprised 0.22%, gluten-free malt 0.01%, and smoked malt 0.045% (median values, Figure 3). Considering these data, it is interesting to look at the beer categories where the brewers mostly employ these malts (grist composition in percentage, Figure 3). For acidulated malt, the Historical Beer (2.23% of grist composition), European Sour Ale (2.13%), American Wild Ale (1.94%), and Amber Malty European Lager (1.17%) were the most used beer categories (Figure 3 and Table S3). In turn, gluten-free malt was most commonly employed in American Wild Ale (3.21%; Figure 3); however, data also show that gluten-free malts were not employed in many different beer categories, such as Smoked Beer, Amber Malty European Lager, and Strong European Beer, among others (Figure 3 and Table S3). Many factors, including low FAN content, poor endosperm enzymatic conversion, and poor foam stability, can explain the restricted use of gluten-free malts (Yang & Gao, 2022). Additionally, some gluten-free malts require specific mashing regimes such as decoction, extrusion, and exogenous enzymes (Yang & Gao, 2022), which impair the usability of gluten-free malts in brewing. Finally, smoked malts were primarily employed in the smoked beer category (15.5%; Figure 3) and amber malty European lager (4.39%; Figure 3).

The second group was composed of both the largest average number of grains used and the percentage of grist composition independent of beer categories, which were the base ($M = 1.30$ grains by recipe), crystal ($M = 0.74$), roasted ($M = 0.43$), and specialty malts ($M = 0.14$) (Figure 3). This result could be interpreted by the fact that brewers will likely use one or two different base malts and zero to one for crystal, roasted, and specialty malts in a recipe, depending on the beer category and style. Considering the percentage of grist composition, the base malt was the most significant fraction, corresponding to 88.68% (median value, Figure 3). This result was expected because base malts contribute to the enzymes needed to convert malt endosperm to fermentable components in wort (Prado et al., 2021). In comparison, the crystal was 5.52%, roasted was 1.85%, and specialty was 2.50% of the grist composition (median values, Figure 3). Based on the grist composition of beer category recipes, base malt comprised 95.54% of the grist for Pale Bitter European Beer, to 62.28% for Smoked Beer, while crystal malt was observed to make up 17.64% of Wood Beer's grist to 1.26% in Historical Beer (Figure 3 and Table S3). It was also observed that Dark British

Beer had a maximum of 13.81%, composed of roasted malts, and 0.27% for Pale Bitter European Beer (Figure 3 and Table S3). Finally, specialty malts comprised 21.76% of Amber Malty European Lager's grist, to 0.74% of Alternative Fermentables Beer (Figure 3 and Table S3).

Historically, different malt-type combinations (grist) in a determined beer category and their associated styles are linked to the expected aroma, flavor, mouthfeel, foam stability, and color that a brewer wants to imprint on it (Stewart et al., 2023). In this sense, the grist composition of a beer category depends on the availability of specific barley cultivars and malting practices, which can be traced to historical beer schools (e.g., English, German, Belgian, and American), directly impacting the sensory aspect of a beer. For example, the malty descriptor found in different beer categories in English brewing schools is linked to heirloom malt varieties developed in the United Kingdom, such as Maris Otter (Morrissy et al., 2022). Not surprisingly, it has been reported that hundreds of aroma and non-volatile compounds are differentially produced by barley varieties from distinct geographical regions, impacting the flavor ("terroir") of a beer (Stewart et al., 2023). Additionally, genetic traits found in barley varieties have downstream effects on malting quality, beer flavor, and abundance of volatile metabolites (Sayre-Chavez et al., 2022). Finally, special malting techniques, such as malt roasting, smoking, or caramelization, bring hundreds of different aromas and nonvolatile compounds into the wort and, consequently, to the beer (Prado et al., 2021).

*3.3. Analysis of hop usage techniques employed in different beer categories*

In addition to malt, hops also play a strong role in defining beer categories and styles. The inflorescence of hops, known as "hop cones," contains lupulin glands composed of essential oils and soft and hard resins (Taniguchi et al., 2014). The soft resin contributes with $\alpha$- and $\beta$-acids to the wort, where $\alpha$-acids are converted to isomerized $\alpha$-acids (iso-$\alpha$-acids) during wort boiling. The presence of iso-$\alpha$-acids impacts bitterness, beer foam stability, and confers microbiological stability (Caballero et al., 2012; Van Holle et al., 2021); however, other hop-derived compounds generated during wort boil or hop/beer storage (e.g., polyphenols, $\beta$-acids transformation products, and oxidative polar compounds) can impact the bitterness perception (Kishimoto et al., 2022; Schönberger & Kostelecky, 2011). International bitterness units, or IBU, determine the bitterness of wort or beer. This spectrophotometric analysis evaluates the concentration of $\alpha$-acids, iso-$\alpha$-acids, $\beta$-acids, and oxidative polar compounds extracted using 2,2,4-trimethylpentane and measured at 275 nm (Kishimoto et al., 2022). Interestingly, the measurement of IBU did not correlate with sensory bitterness intensity in a linear form, and beers with high residual extract and ethanol have high IBU values despite being balanced, as observed for some beer categories and styles, like stouts, APAs, and IPAs (Hahn et al., 2018). The data gathered in this work regarding cold- and hot-fermenting beer categories (Figures 4A and B) showed that two early hop additions are the major contributors to IBU, which are boil and first wort hopping (FWH) techniques (Figures 4A and B). For cold fermenting beer, the FWH contributes 15.99 IBUs, and boil 10.62 IBUs (median values, Figure 4A), while in hot fermenting beers, the FWH contributes 18.67 IBUs and boil 12.75 IBUs (median values, Figure 4B). The use of hops at the beginning of wort boiling is a standard technique for promoting the formation of iso-$\alpha$-acids in wort (Rutnik et al., 2023). The percentized number of recipes that employed boil hopping varied from 87% (Smoked Beer category) to 100% for cold-fermented beers (Pale Bitter European Beer) (Figure 4A) and 51% (Alternative Fermentables Beer) to 100% (IPA) for hot-fermented beers (Figure 4B). However, despite the historical and widespread use of boil hopping, physicochemical factors (e.g., pH, SG, boiling temperature, boiling duration) differentially impact the rate of $\alpha$-acids isomerization (Jaskula et al., 2008). Also, boil hopping promotes the "kettle hop aroma" of using noble hops rich in humulene (Peacock & Deinzer, 1981), which is oxidized during boiling or hop storage and generates humulinone, imparting bitterness to the beer (Rutnik et al., 2023).

Another technique to add bitterness to beer is first wort hopping (FWH), where hops are added to the wort after run-off and before boiling. The central idea behind using FWH is to increase the solubility of bittering compounds when using hop cones instead of pellets because of the slow release of bittering compounds from the first (Forster et al., 2022). This technique was employed at the end of the 19th century and revived at the beginning of the 21st century as an alternative to boil hopping, where it was described that FWH increases bitterness in wort without being harsh (Forster et al., 2022). The percentage of recipes that employed FWH varied from 34% (Smoked Beer category) to 84% (Pale Bitter European Beer) for cold-fermented beers (Figure 4A) and 17% (Alternative Fermentables and Wood Beer categories) to 94% (IPA) for hot-fermented beers (Figure 4B). These data indicate that the FWH technique is more widespread among brewers than was initially thought. Finally, mash hopping is also an early addition technique where hops are added during the wort mash-in phase to aggregate essential oils instead of iso-$\alpha$-acids since the mashing temperature is below that necessary to convert $\alpha$-acids to iso-$\alpha$-acids (Torgals, 2015). The contribution of mash hopping for different cold-fermented beer categories was 1.45 IBUs (median value, Figure 4A) and 4.70 IBUs for hot-fermented beers (median value, Figure 4B). The data analysis of beer recipes shows that the mash hopping procedure was employed in 29% of recipes in the Pale Bitter European Beer category (Figure 4A). In comparison, 79% of IPA recipes also used this hopping technique (Figure 4B).

In turn, late hop additions mainly contribute essential oils and volatile molecules to the final beer (Passaghe et al., 2024), and different late hop techniques have been developed, such as dry hopping, dry hopping during high krausen (dry hop HK), hopback, and hop stand, among others. Recipe data analysis indicated that dry hop HK and hop stand are employed less in cold and hot fermented beers than dry hopping (Figures 4A and B), except for Amber Bitter European Lager, Pale Bitter European Lager, IPA, and Pale American Ale (Figures 4A and B). In contrast, dry hopping appears to be extensively employed in cold and hot fermented beers (Figures 4A and B).

Finally, analysis of the relative bitterness ratio (RBR) of cold and hot fermented beers did not indicate any statistical differences among the recipes (Figures 4A and B).

*3.4. Self-organizing maps of beer categories and styles with a proposition of a new classification*

Considering the average number of different malt subtypes and hops employed by different beer styles, malt types, and beer hopping methods, as well as the average values of SG, final gravity (FG), ADF, ABV, SRM, and IBU for each beer style, a dissimilarity matrix was generated to classify each beer style into major superclusters and clusters using the self-organizing map (SOM) technique (Figures 5–8; Table 1).

Four superclusters containing five to six clusters were observed in the SOM analysis (Figures 5–8; Table 1). Supercluster 1 represents all cold-fermented beer styles (Figure 5 and Table 1). In contrast, superclusters 2–4 represent all hot-fermented beers, indicating a clear distinction between malt and hop usage and style data. Moreover, the phylum and class names were attributed to each supercluster and cluster (Table 1).

Supercluster 1 (phylum "Cold Fermented Classics"; Figure 5) is composed of six classes (dark lager, smoke and specialty lager, strong dark lager, amber lager, nest lager, pale, and premium Lagers) (Table 1). The SOM data indicated that clusters 6 (Amber Lagers; Table 1) and 11 (Pale and Premium Lagers; Table 1) shared similarities in terms of malt and hop usage and style data (Figure 5), while Cluster 1 (Dark Lagers; Table 1), 2 (Smoked and Specialty Lagers; Table 1), 3 (Strong Dark Lagers; Table 1), and 7 (Märzen; Table 1) were the most similar. Interestingly, Märzen was the only beer style represented in Cluster 7 (Figure 5). Märzen beers were considered one of the first larger styles brewed in Germany (Hutzler et al., 2023).

Supercluster 2 (phylum "Malty and Roasted Ales"; Figure 6) comprises six clusters (Brown Ales, Stouts and Porters, Malty Ales, Light Malty Ales, Historical Specialty and Seasonal Ales, Strong Adjunct Beers) (Table 1) that share high values of SRM, final gravity, and malt complexity, with a predominance of crystal, roasted, and specialty malts. Clusters 4 (Brown Ales; Table 1) and 5 (Stouts and Porters; Table 1) were closely related, likely because of the presence of dark roasted grains and moderate hopping (Figure 6). Cluster 9 (Malty Ales; Table 1) includes seasonal and traditional UK ales such as Scottish Export and Irish Red Ale, known for their malt-forward profiles. Cluster 13 (Historical and Seasonal Ales; Table 1) groups several historically significant styles, such as Bière de Garde and Oud Bruin, which exhibit high malt complexity and sourness. This supercluster reflects traditional European brewing heritage, where malt diversity and fermentation nuances are central to style identity.

Supercluster 3 (phylum "Diverse Fermentation Profiles"; Figure 7 and Table 1) includes six clusters (Smoked Wheat Beers, Light and Easy Beers, Sugar and Grain Beers, Sour and Wild Beers, Alternative and Specialty Grain Beers, and Belgian Style Ales), unified by low-to-moderate SRM and ABV, and high variability in malt and hop profiles. Cluster 16 (Smoked Wheat Beers; Table 1) was represented solely by Piwo Grodziskie, a unique Polish wheat beer brewed with oak-smoked malts. Cluster 17 (Light and Easy Beers; Table 1) encompasses approachable styles, such as Cream Ale and Blonde Ale, with low bitterness and restrained fermentation profiles. Clusters 21 and 22 stand out for including sour and wild fermented beers (e.g., Lambic, Gose, and Saison; Table 1), where adjuncts such as fruits and alternative grains play a significant role, highlighting non-standardized fermentation routes. Finally, Cluster 23 (Belgian Style Ales; Table 1) reflects Belgian yeasts' aromatic complexity and high ester production and their use of sugar adjuncts for dry and high-ABV profiles (Postigo et al., 2024).

Supercluster 4 (phylum "Hop-Forward and Strong Ales"; Figure 8 and Table 1) encompasses five clusters (Dark Strong Ales, Amber and Pale Ales, Pale Strong Ales, Hoppy and Clone Beers, New IPAs) that show higher IBU, ABV, and hop complexity, especially from late hop additions such as dry hopping and hop stands. Cluster 15 (Dark Strong Ales; Table 1) includes robust and aged beer styles, such as Imperial Stout and English Barleywine, where high hopping and malt intensity coexist (Caballero et al., 2012; Hahn et al., 2018). Clusters 19 and 20 aggregated traditional British and American ales, respectively, with strong hop backbones and balanced malt profiles (Table 1). Cluster 24 (Hoppy and Clone Beers) and Cluster 25 (New IPAs) displayed the highest levels of hop diversity and usage, especially in whirlpool and dry-hopping steps, which is typical for American and New England IPAs (Table 1). This supercluster represents the modern brewing trend toward aroma-rich and hop-intense beers.

## 4. Conclusion

This study presents a comprehensive and data-driven classification system for beer categories and styles that integrates over 62,000 curated recipes and employs self-organizing maps (SOMs) for unsupervised learning. Our results revealed four major superclusters: Cold Fermented Classics, Malty and Roasted Ales, Diverse Fermentation Profiles, and Hop-Forward and Strong Ales, each characterized by unique patterns in malt and hop usage, fermentation parameters, and sensory attributes. This novel approach reflects the historical and technological evolution of brewing practices and provides a robust framework for understanding beer diversity beyond traditional classifications. The proposed taxonomy is rooted in objective recipe data and supports brewers, researchers, and sensory analysis to navigate the complex modern brewing landscape. Furthermore, the framework encourages future investigations linking biochemical profiles, consumer preferences, and brewing innovations.

**Funding:** This work was supported by "CNPq/MCTI/CT-BIOTEC N° 30/2022 - Linha 2: Novas tecnologias em Biotecnologia" from Conselho Nacional de Desenvolvimento

Científico e Tecnológico – CNPq [grant number 440226/2022-8], by "Programa Pesquisador Gaúcho - PqG" from Fundação de Amparo à Pesquisa do Estado do Rio Grande do Sul - FAPERGS [Edital FAPERGS 07/2021 and grant number 21/2551-0001958-1], by "INOVA CLUSTERS TECNOLÓGICOS" from Fundação de Amparo à Pesquisa do Estado do Rio Grande do Sul - FAPERGS [Edital FAPERGS 02/2022 and grant number 22/2551-0000834-8] and by "Programa de Redes Inovadoras de Tecnologias Estratégicas do Rio Grande Do Sul – RITEs-RS" from Fundação de Amparo à Pesquisa do Estado do Rio Grande do Sul - FAPERGS [Edital FAPERGS 06/2021 and grant number 22/2551-0000397-4]. The sponsors had no role in the study design, collection, analysis, interpretation of data, writing of the report, or the decision to submit the article for publication.

**Declarations of interest:** None

**Human and animal rights:** The experiments included in this study did not involve any animal or human participants.

**Declaration of generative AI and AI-assisted technologies in the writing process**

During the preparation of this study, ChatGPT (https://chatgpt.com) and PaperPal (https://paperpal.com) were used to improve the readability and language of the manuscript. After using these tools and services, the author reviewed and edited the content as needed, and took full responsibility for the content of the published article.

& Van Landschoot, A. (2021). Relevance of hop terroir for beer flavour. *Journal of the Institute of Brewing*, *127*(3), 238–247. https://doi.org/10.1002/jib.648

Villacreces, S., Blanco, C. A., & Caballero, I. (2022). Developments and characteristics of craft beer production processes. *Food Bioscience*, *45*, 101495. https://doi.org/10.1016/j.fbio.2021.101495

Yang, D., & Gao, X. (2022). Progress of the use of alternatives to malt in the production of gluten-free beer. *Critical Reviews in Food Science and Nutrition*, *62*(10), 2820–2835.
30

**Tables**

**Table 1.**

| Superclusters | Phylum | Clusters | Class | Styles |
|---|---|---|---|---|
| Supercluster 1 | Cold Fermented Classics | Cluster 1 | Dark Lagers | Altbier, Czech Dark Lager, International Dark Lager, Schwarzbier |
| Supercluster 1 | Cold Fermented Classics | Cluster 2 | Smoked and Specialty Lagers | Classic Style Smoked Beer, Munich Dunkel, Rauchbier, Specialty Smoked Beer |
| Supercluster 1 | Cold Fermented Classics | Cluster 3 | Strong Dark Lagers | Baltic Porter, Doppelbock, Dunkles Bock, Eisbock |
| Supercluster 1 | Cold Fermented Classics | Cluster 6 | Amber Lagers | Czech Amber Lager, Helles Bock, International Amber Lager, Kellerbier: Amber Kellerbier, Vienna Lager |
| Supercluster 1 | Cold Fermented Classics | Cluster 7 | Fest Lagers | Märzen |
| Supercluster 1 | Cold Fermented Classics | Cluster 11 | Pale and Premium Lagers | Czech Pale Lager, Czech Premium Pale Lager, Festbier, German Helles Exportbier, German Leichtbier, German Pils, International Pale Lager, |

| Superclusters | Phylum | Clusters | Class | Styles |
|---|---|---|---|---|
| | | | | Kellerbier: Pale Kellerbier, Kölsch, Munich Helles |
| Supercluster 2 | Malty and Roasted Ales | Cluster 4 | Brown Ales | American Brown Ale, British Brown Ale, Dark Mild |
| Supercluster 2 | Malty and Roasted Ales | Cluster 5 | Stouts and Porters | American Porter, American Stout, English Porter, Foreign Extra Stout, Irish Extra Stout, Irish Stout, London Brown Ale, Oatmeal Stout, Sweet Stout, Tropical Stout |
| Supercluster 2 | Malty and Roasted Ales | Cluster 9 | Malty Ales | Autumn Seasonal Beer, Belgian Dubbel, Irish Red Ale, Scottish Export, Scottish Heavy, Wee Heavy, Winter Seasonal Beer |
| Supercluster 2 | Malty and Roasted Ales | Cluster 8 | Light Malty Ales | Scottish Light |
| Supercluster 2 | Malty and Roasted Ales | Cluster 13 | Historical Specialty and Seasonal Ales | Bière de Garde, Dunkles Weissbier, Flanders Red Ale, Kentucky Common, Oud Bruin, Pre-Prohibition Porter, Roggenbier |

| Superclusters | Phylum | Clusters | Class | Styles |
| --- | --- | --- | --- | --- |
| Supercluster 2 | Malty and Roasted Ales | Cluster 14 | Strong Adjunct Beers | Spice, Herb, or Vegetable Beer, Weizenbock |
| Supercluster 3 | Diverse Fermentation Profiles | Cluster 16 | Smoked Wheat and Oat Beers | Piwo Grodziskie |
| Supercluster 3 | Diverse Fermentation Profiles | Cluster 17 | Light and Easy Beers | American Lager, Blonde Ale, Cream Ale, Pre-Prohibition Lager, Trappist Single |
| Supercluster 3 | Diverse Fermentation Profiles | Cluster 18 | Sugar and Grain Beers | Alternative Sugar Beer, Fruit and Spice Beer, Sahti |
| Supercluster 3 | Diverse Fermentation Profiles | Cluster 21 | Sour and Wild Beers | Berliner Weisse, Fruit Lambic, Gose, Gueuze, Lambic, Lichtenhainer, Mixed-Fermentation Sour Beer, Weissbier, Wild Specialty Beer, Witbier |
| Supercluster 3 | Diverse Fermentation Profiles | Cluster 22 | Alternative and Specialty Grain Beers | Alternative Grain Beer, American Wheat Beer, Australian Sparkling Ale, Brett Beer, Fruit Beer, Saison |

| Superclusters | Phylum | Clusters | Class | Styles |
|---|---|---|---|---|
| Supercluster 3 | Diverse Fermentation Profiles | Cluster 23 | Belgian Style Ales | American Light Lager, Belgian Blond Ale, Belgian Golden Strong Ale, Belgian Pale Ale, Belgian Tripel, British Golden Ale, Experimental Beer, Mixed-Style Beer, Specialty Fruit Beer |
| Supercluster 4 | Hop-Forward and Strong Ales | Cluster 15 | Dark Strong Ales | Belgian Dark Strong Ale, English Barleywine, Imperial Stout, Old Ale, Specialty Wood-Aged Beer, Wood-Aged Beer |
| Supercluster 4 | Hop-Forward and Strong Ales | Cluster 19 | Amber and Pale Ales | American Amber Ale, Best Bitter, Burton Ale, California Common, Ordinary Bitter, Strong Bitter |
| Supercluster 4 | Hop-Forward and Strong Ales | Cluster 20 | Pale Strong Ales | American Barleywine, American Strong Ale, British Strong Ale, Wheatwine |
| Supercluster 4 | Hop-Forward and Strong Ales | Cluster 24 | Hoppy and Clone Beers | American Pale Ale, Clone Beer, English IPA |

| Superclusters | Phylum | Clusters | Class | Styles |
|---|---|---|---|---|
| Supercluster 4 | Hop-Forward and Strong Ales | Cluster 25 | New IPAs | American IPA, Double IPA, Specialty IPA: Belgian IPA, Specialty IPA: Black IPA, Specialty IPA: Brown IPA, Specialty IPA: New England IPA, Specialty IPA: Red IPA, Specialty IPA: Rye IPA, Specialty IPA: White IPA |

**Figure legends**

**Figure 1.** Dot plots showing the distribution and average number of ingredients used in beer recipes classified as (A) cold-fermented and (B) hot-fermented. Categories include grains (malt types and subtypes), hops, dry and liquid malt extracts, adjuncts (e.g., non-malted grains), sugars (e.g., honey, caramelized sugars), and fruits. Statistical differences were evaluated using Bootstrap-t methods for means and represented with trimmed average (μ) and 95% confidence intervals. The dashed line represents the average number of components employed in cold- and hot-fermented beers.

**Figure 2.** Heatmap comparing the number of unique ingredients used in different beer categories (numbers indicated inside the cells), including grains (malt types and subtypes), hops, dry and liquid malt extracts, adjuncts (e.g., non-malted grains), sugars (e.g., honey, caramelized sugars), and fruits in hot- and cold-fermented beers (A). In (B), statistical evaluation of different ingredients employed in cold- and hot-fermented beers categories. Statistical data analysis is indicated in the graphics.

**Figure 3.** Heatmap showing the percentize of beer recipes that employed a specific malt type (base, crystal, roasted, specialty, acidulated, smoked, and gluten-free). The percentize numbers are indicated within each cell. A gray gradient color, from light gray to dark gray, was added to each cell. In addition, the average number of different malt types per recipe and grist composition percentage for each malt type across beer categories are indicated by the boxplot graphics at the top of the heatmap. Finally, the average number of different malt types per recipe and grist composition percentage for each beer category are indicated by the barplots on the right side of the heatmap. The colors of barplots indicated the different malt types employed in each beer category (indicated by the legend in the lower right).

**Figure 4.** Heatmaps representing (A) cold-fermented and (B) hot-fermented beer categories, showing the percentized number of recipes and the average International Bitterness Units (IBU) contributions for each hop addition method, including boil, first wort hopping (FWH), mash, whirlpool, hop stand, hopback, dry hop, and dry hop at high krausen. The box plot at the top of the heatmap compares each hopping method's contribution to the total IBU. Additionally, the relative bitterness ratio (RBR) in each beer category is indicated by the box plot on the right side of the heatmap.

**Figure 5.** Self-organizing map (SOM) of cold-fermented beer styles (Supercluster 1).

The SOM output shows clustering of cold-fermented beer styles into six distinct clusters (1, 2, 3, 6, 7, and 11). Clusters are defined by malt and hop composition and style-specific brewing parameters. The numbers within each cell indicate the average number of different malts and types for each malt and hop beer style. Considering the style-specific brewing parameters, the numbers inside the cell indicate the average value for each beer style.

**Figure 6.** Self-organizing map (SOM) of malty and roasted hot-fermented ales (Supercluster 2). SOM output showing six clusters representing hot-fermented beer styles characterized by malt-forward profiles (4, 5, 8, 9,13, and 14). Clusters are defined by malt and hop composition and style-specific brewing parameters. The numbers within each cell indicate the average number of different malts and types by each beer style for malt and hop. Considering style-specific brewing parameters, the numbers inside the cell indicated the average value for each beer style.

**Figure 7.** Self-organizing map (SOM) of diverse hot-fermentation profiles (Supercluster 3).

The SOM output highlights six clusters characterized by non-standard fermentation approaches or adjunct usage (16, 17, 18, 21, 22, and 23). Clusters are defined by malt and hop composition and style-specific brewing parameters. The numbers within each cell indicate the average number of different malts and types for each malt and hop beer style. Considering the style-specific brewing parameters, the numbers inside the cell indicate the average value for each beer style.

**Figure 8.** Self-organizing map (SOM) of hop-forward and strong ales (Supercluster 4).

The SOM output of beer styles with pronounced hop characters and/or elevated alcohol content was grouped into five clusters (15, 19, 20, 24, and 25). Clusters are defined by malt and hop composition and style-specific brewing parameters. The numbers within each cell indicate the average number of different malts and types for each malt and hop beer style. Considering the style-specific brewing parameters, the numbers inside the cell indicate the average value for each beer style.

**Figure 1.**

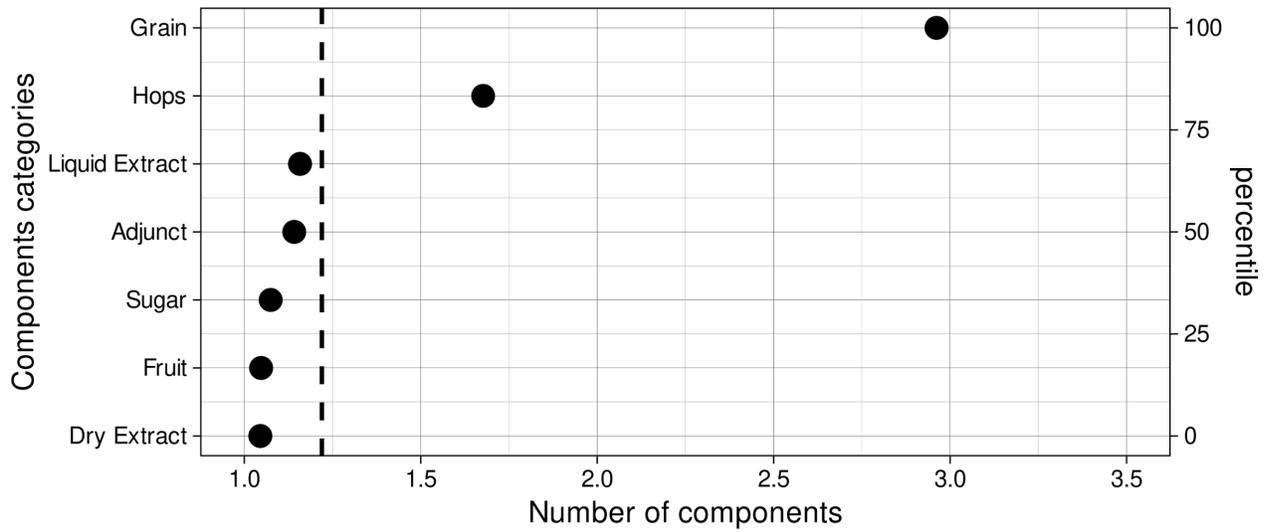

A  Cold fermented

$t_{bootstrapped} = 6.73$, $p = 0.04$, $\widehat{\mu}_{trimmed} = 1.22$, $CI_{95\%}$ [0.25, 2.19], $n_{obs} = 7$

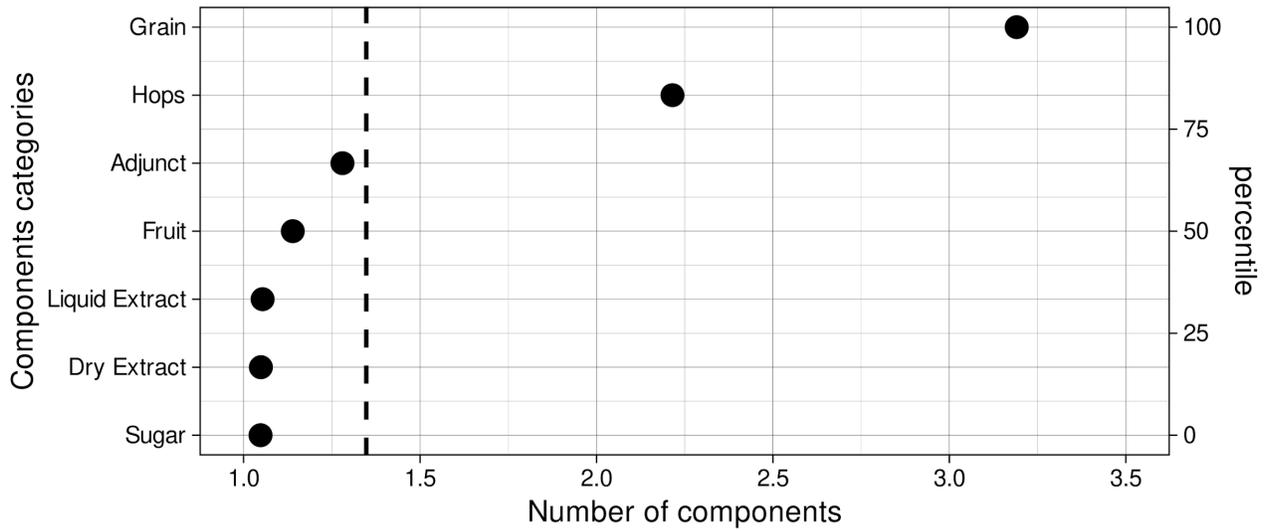

B  Hot fermented

$t_{bootstrapped} = 3.94$, $p = 0.13$, $\widehat{\mu}_{trimmed} = 1.35$, $CI_{95\%}$ [-2.14, 4.84], $n_{obs} = 7$

Figure 2.

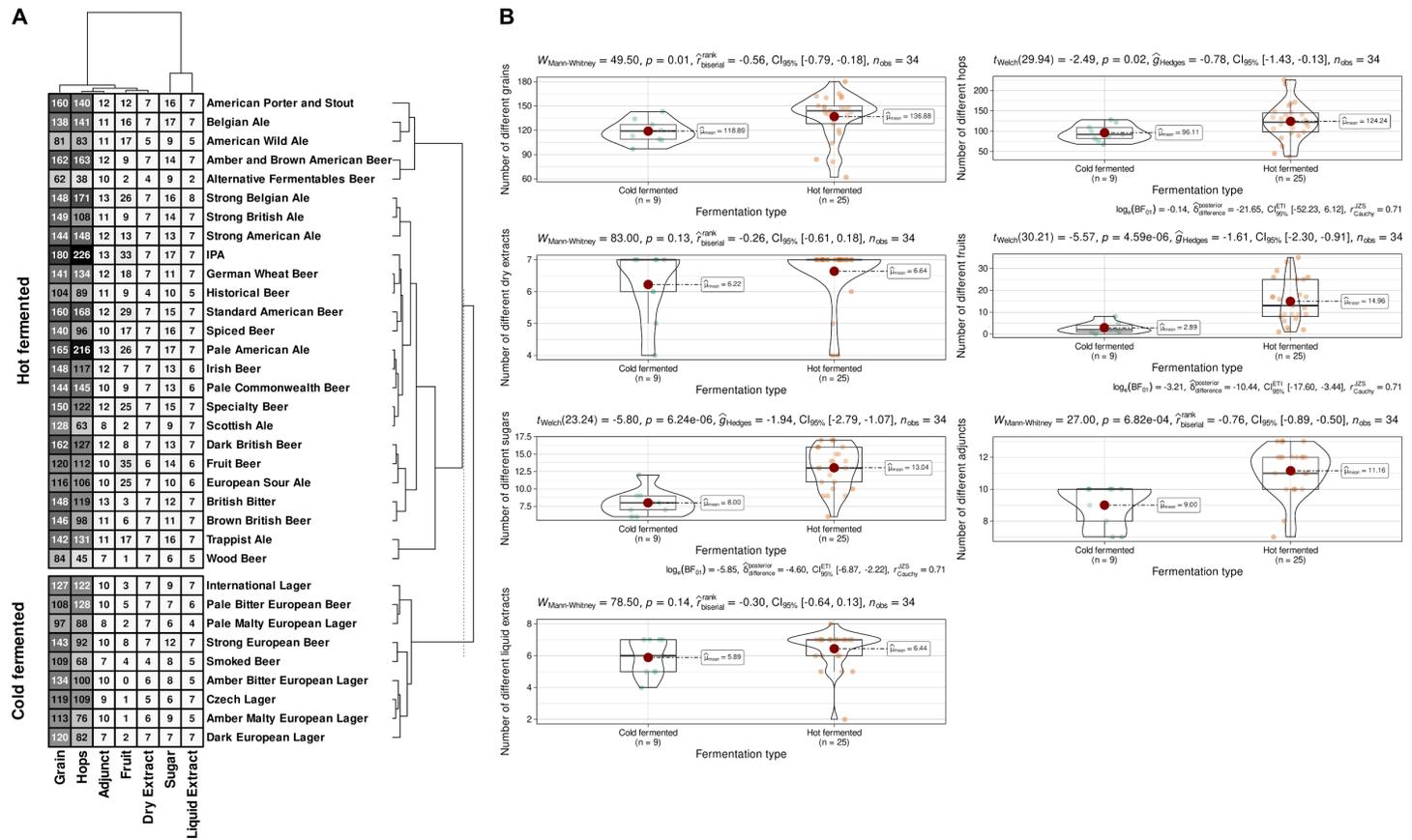

**Figure 3.**

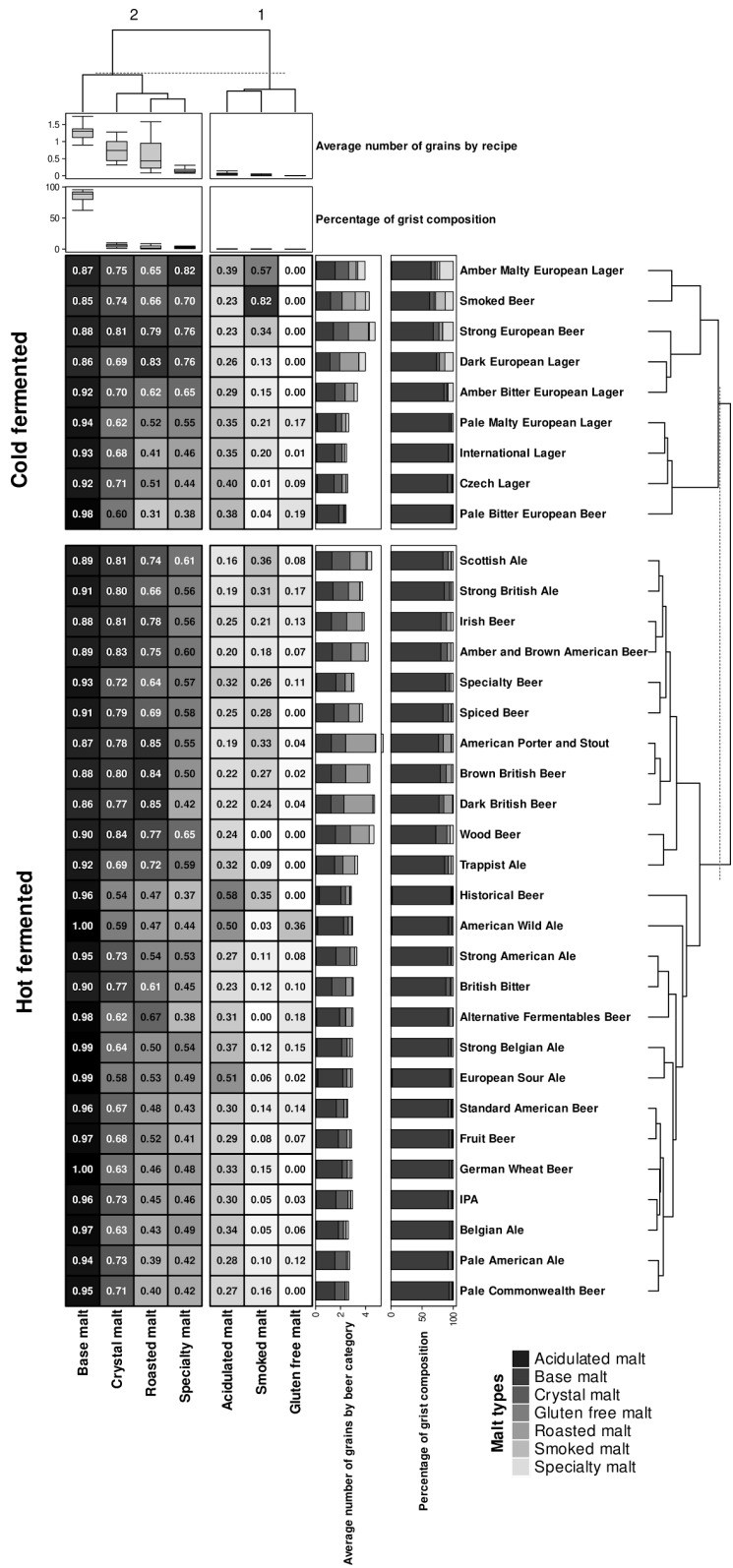

Figure 4.

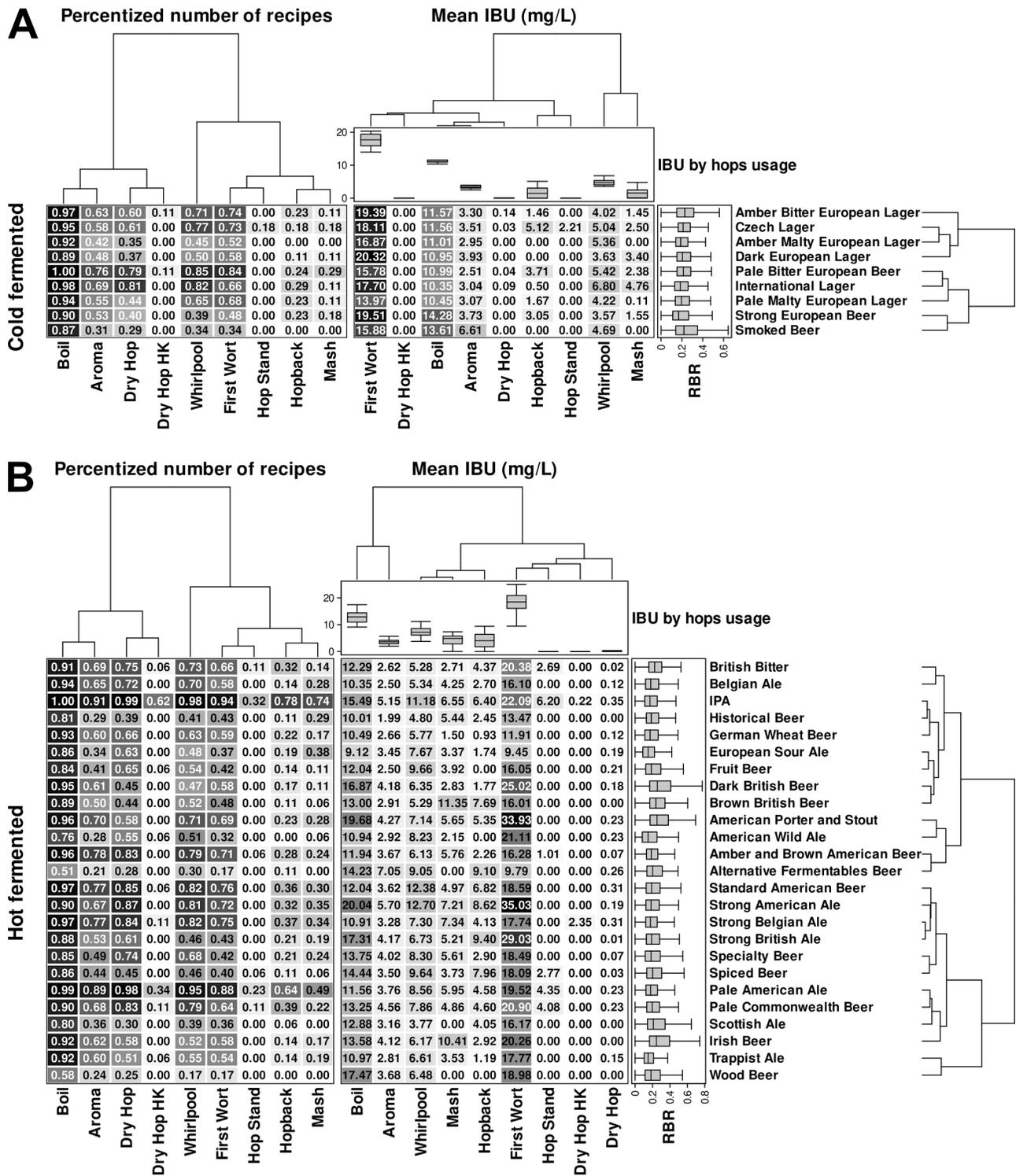

**Figure 5.**

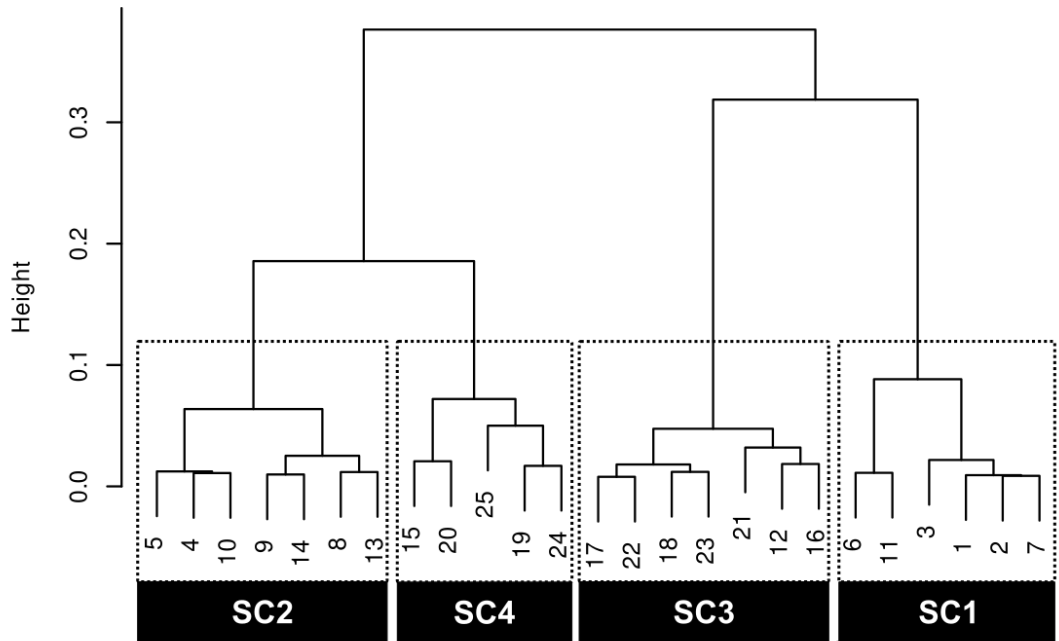

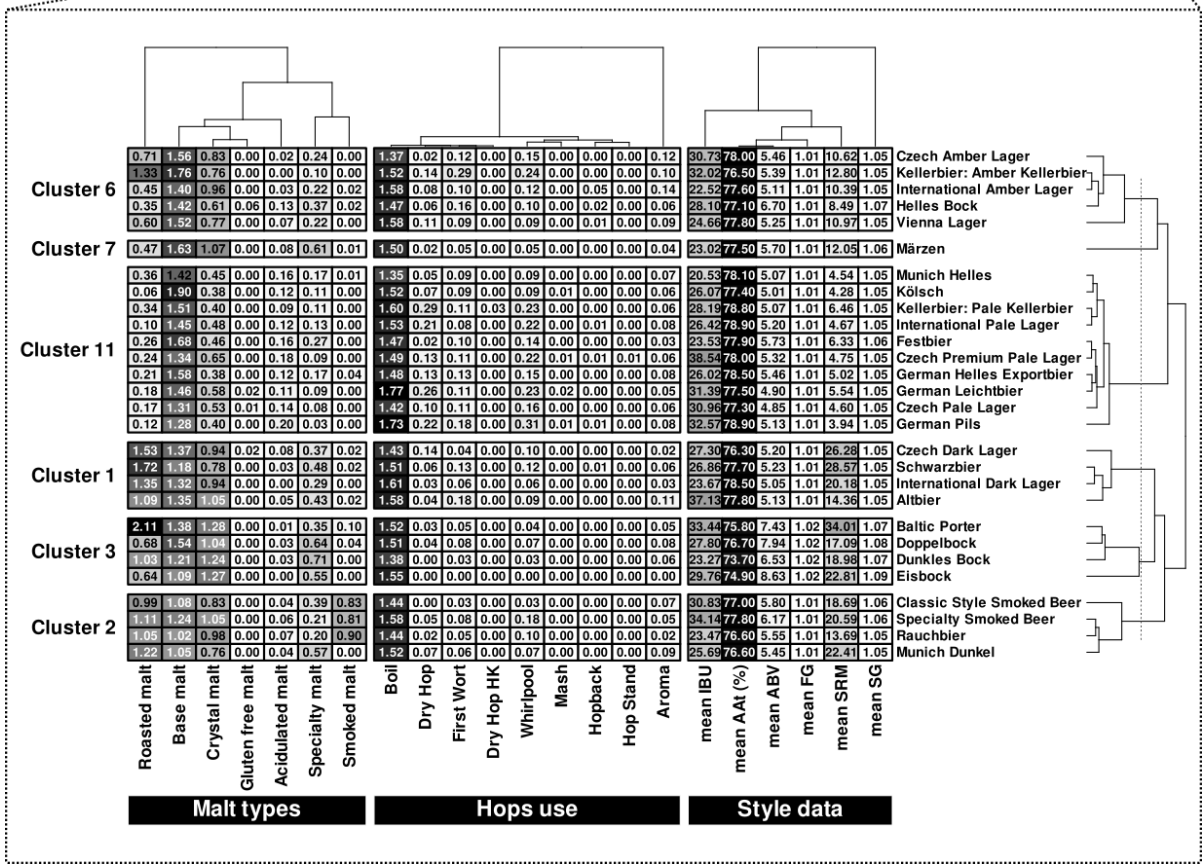

**Figure 6.**

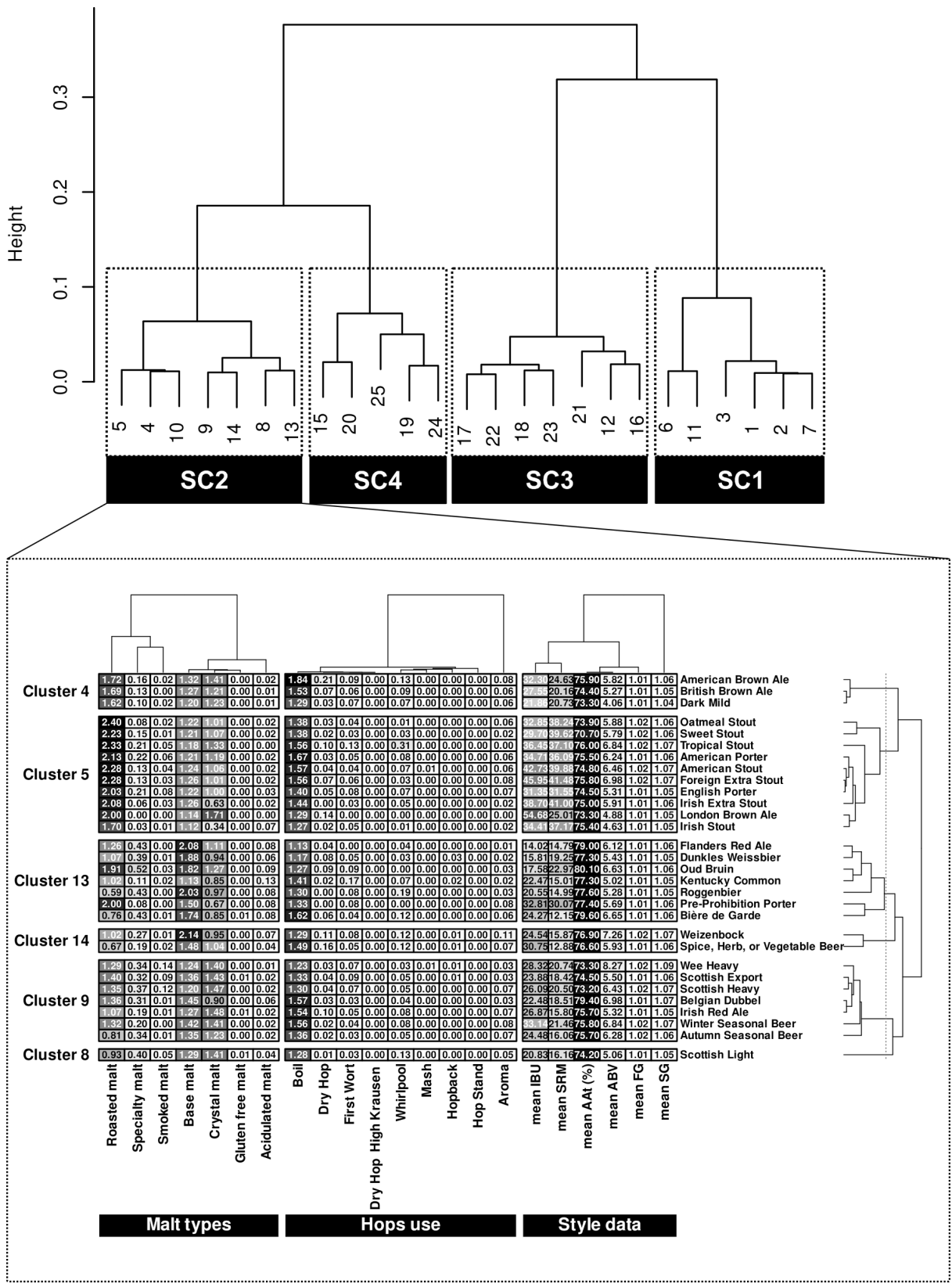

**Figure 7.**

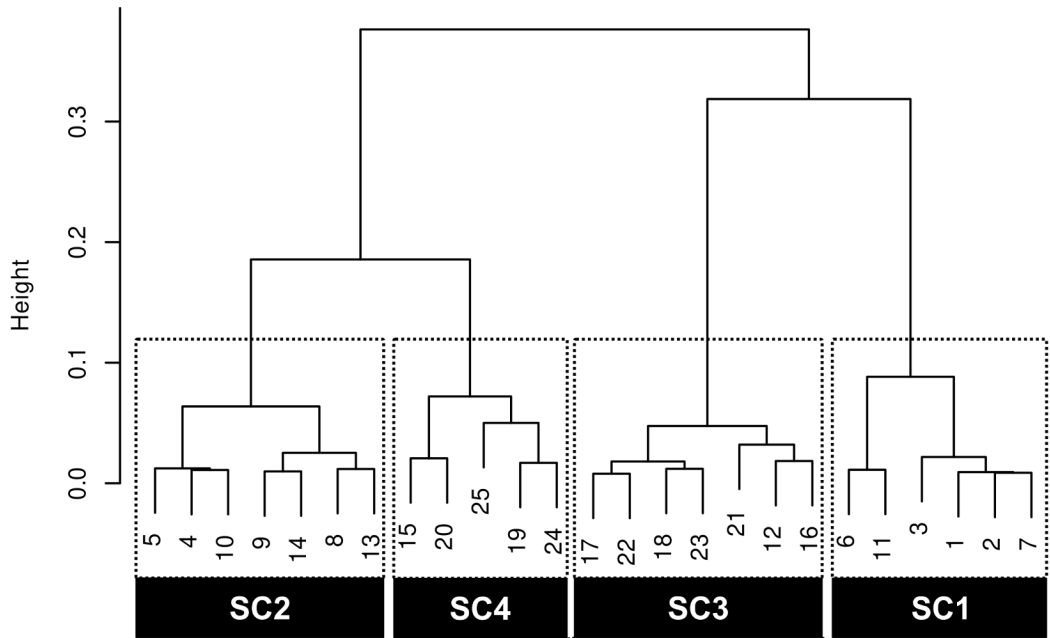
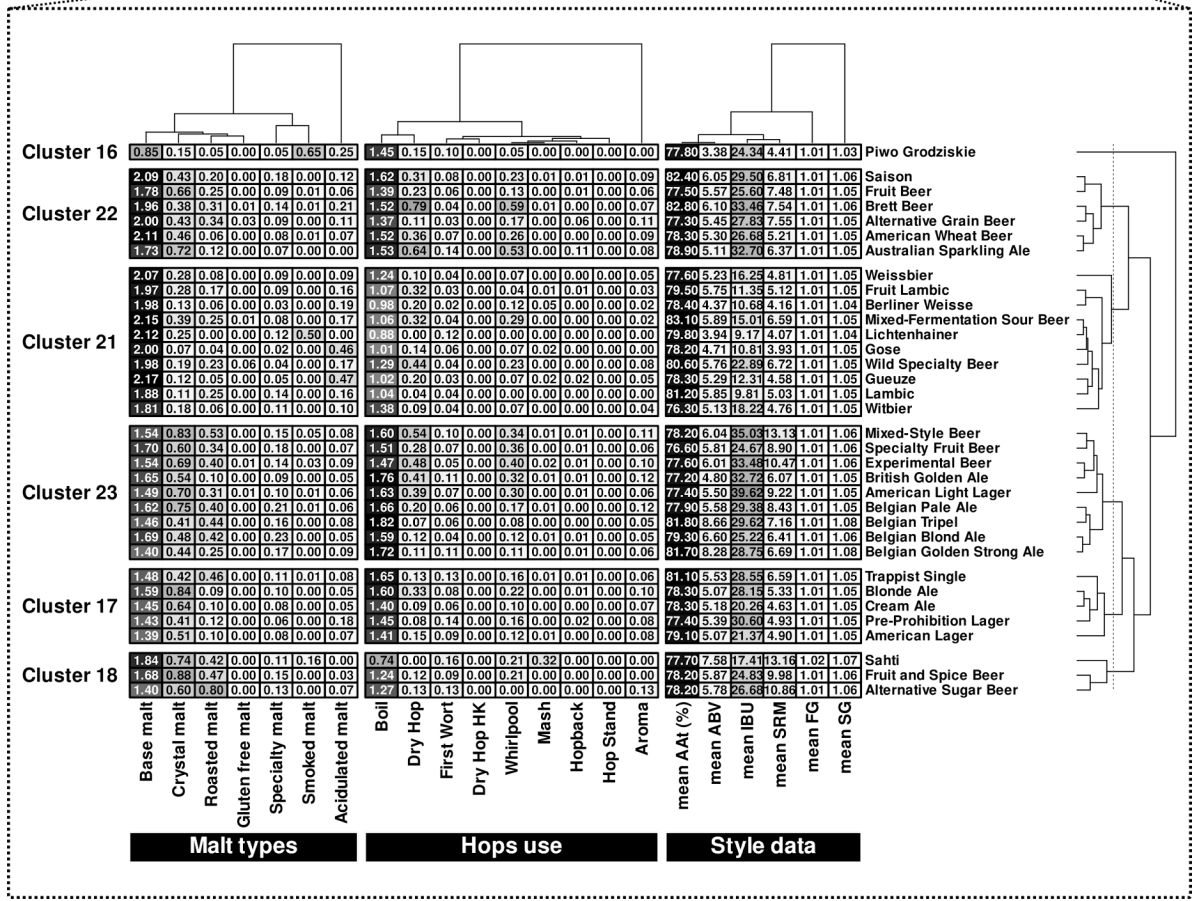

**Figure 8.**

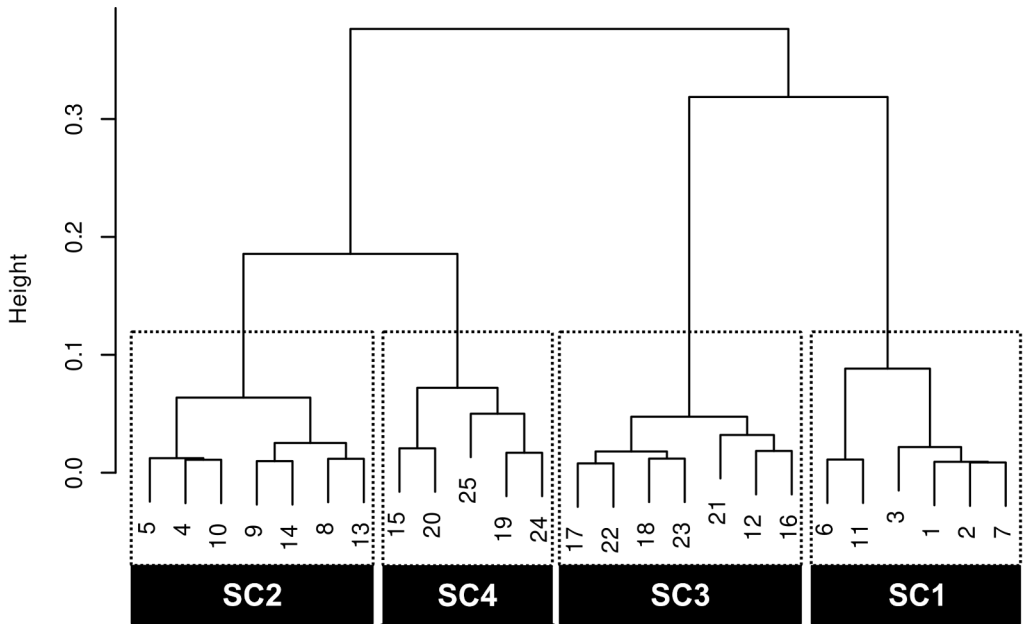

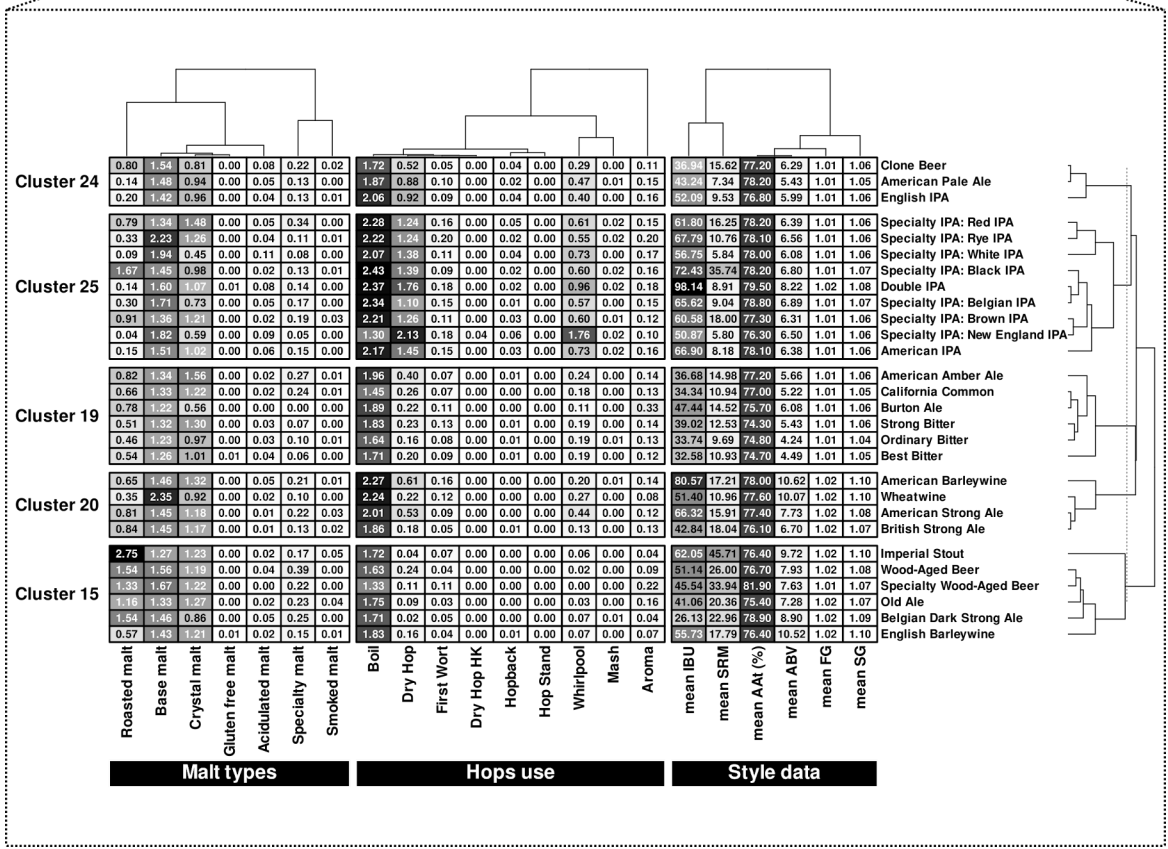